\documentclass{ecai}  % use option [doubleblind] for double blind submission and hiding the authors section

\usepackage{graphicx}
\usepackage{latexsym}

\usepackage{amssymb}
\usepackage{amsmath}
\usepackage{amsthm}
\usepackage{booktabs}
\usepackage{enumitem}
\usepackage{graphicx}
\usepackage{color}

\usepackage{multicol}
\usepackage{multirow}
\usepackage{array}
\usepackage{ulem}
\usepackage{hyperref}
\usepackage{caption}
\usepackage{subcaption}
\begin{document}

\begin{frontmatter}

\title{FltLM: An Intergrated Long-Context Large Language Model for Effective Context Filtering and Understanding}

\author[A]{\fnms{Jingyang}~\snm{Deng}}
\author[B]{\fnms{Zhengyang}~\snm{Shen}\thanks{Co-corresponding Author. Email: shenzhy@pku.edu.cn}}
\author[C]{\fnms{Boyang}~\snm{Wang}} 
\author[B]{\fnms{Lixin}~\snm{Su}} 
\author[B]{\fnms{Suqi}~\snm{Cheng}} 
\author[B]{\fnms{Ying}~\snm{Nie}} 
\author[B]{\fnms{Junfeng}~\snm{Wang}} 
\author[B]{\fnms{Dawei}~\snm{Yin}} 
\author[A]{\fnms{Jinwen}~\snm{Ma}\thanks{Co-corresponding Author. Email: jwma@math.pku.edu.cn}}

\address[A]{School of Mathematical Sciences and LMAM, Peking University}
\address[B]{Baidu, Inc.}
\address[C]{CCSE, Beihang University}

\begin{abstract}
The development of Long-Context Large Language Models (LLMs) has markedly advanced natural language processing by facilitating the process of textual data across long documents and multiple corpora. However, Long-Context LLMs still face two critical challenges: The \textit{lost in the middle} phenomenon, where crucial middle-context information is likely to be missed, and the \textit{distraction} issue that the models lose focus due to overly extended contexts. To address these challenges, we propose the Context Filtering Language Model (FltLM), a novel integrated Long-Context LLM which enhances the ability of the model on multi-document question-answering (QA) tasks. Specifically, FltLM innovatively incorporates a context filter with a soft mask mechanism, identifying and dynamically excluding irrelevant content to concentrate on pertinent information for better comprehension and reasoning. Our approach not only mitigates these two challenges, but also enables the model to operate conveniently in a single forward pass. Experimental results demonstrate that FltLM significantly outperforms supervised fine-tuning and retrieval-based methods in complex QA scenarios, suggesting a promising solution for more accurate and reliable long-context natural language understanding applications.
\end{abstract}

\end{frontmatter}

\section{Introduction}
\label{intro}

The advent of Long-Context Large Language Models (LLMs) marks a significant advancement in natural language processing, addressing the increasing demand for comprehending and generating extensive textual data. The need for such models arises from the vast amount of information contained in lengthy documents and multiple corpora, which general LLMs often struggle to process. Long-context LLMs promise to revolutionize a range of applications, including in-depth cross-document question answering \cite{bai2023longbench}, comprehensive document summarization \cite{chang2023booookscore} and sophisticated content generation  \cite{han2023lm,zhang2024soaring}, exhibiting a promising future with huge development potential.

Current research has introduced various strategies to extend context window of LLMs, ranging from modifications in positional encoding during the continual pre-training stage, such as Positional Interpolation (PI) \cite{chen2023extending}, NTK-aware interpolation, and YaRN \cite{peng2023yarn}, to efficient training methods such as LongLoRA \cite{chen2023longlora}, LongQLoRA \cite{yang2023longqlora}, and PoSE \cite{zhu2023pose}. Furthermore, to address the computational challenges posed by the quadratic complexity of the self-attention mechanism, distributed training approaches have been developed. These techniques, such as sequence parallelism \cite{li2023sequence} and distributed attention mechanisms, enable scaling up both model size and the context window length, thereby enhancing the processing capacity of LLMs.

With the help of these advancements, the research community has witnessed the emergence and open-sourcing of numerous Long-Context LLMs. However, the evaluation of these models has mainly focused on their long-context modeling capabilities, with perplexity as a metric, or on some basic information retrieval tasks such as pass-key retrieval \cite{chen2023extending} and the Needle in the Haystack \cite{greg2023needle}. Relative less attention has been paid to enhancing model performance in downstream real-world tasks, while it is crucial for unlocking the full potential of these models. To get further in this field, we mainly focus on a challenging yet ubiquitous multi-hop multi-document question-answering (QA) task, which asks the model to gather and integrate information from multiple pieces of text to generate a correct answer. Unlike single-hop QA tasks, where the answer to a question can be found straightforwardly within a single sentence or document, multi-hop QA involves reasoning across several documents or parts of a context to synthesize the answer.

Unfortunately, Long-Context LLMs face significant challenges in such tasks, as demonstrated by recent research showing that they struggle with intricate long dependency tasks \cite{li2023loogle}, such as QA tasks that requires model to retrieve multiple pieces of information or engage in comprehension and reasoning. We believe that these challenges can be attributed to two main factors:
\begin{itemize}
    \item \textbf{Lost in the middle phenomenon}. From a data perspective, natural language exhibits inherent biases, as people tend to prioritize important information at the beginning and end of contexts. As a result, both pre-training and instruction-tuning data may share such a trend. Under the supervision of the next token prediction task, Long-Context LLMs may also mimic this human tendency, overemphasizing the beginning and end tokens while sometimes ignoring the middle ones. This \textit{lost in the middle} phenomenon was initially observed by Liu et al. \cite{liu2024lost}, who revealed that Long-Context LLMs struggle to seek relevant information when the ground truth document is located in the middle of the input context.

    \item \textbf{Distraction issue}. From a model perspective, as the context window extends, an increasing number of tokens are involved in the self-attention mechanism, where attention scores for each query may dispersed across too many keys. In such a situation, keys with different semantic meaning are more likely to overlap and become hard to distinguish by the query, making it  difficult for the model to focus on relevant information. This \textit{distraction} issue was first proposed by Tworkowski et al. \cite{tworkowski2024focused}.
\end{itemize}

 To address the \textit{lost in the middle} phenomenon, several studies utilize additional information from multi-document QA training data to generate augmented answers \cite{yu2023paraphrasing, junqing2023never}, noticing that it is easy to acquire ground-truth documents when constructing training data. These augmented answers provide more supervision signals and can help the model to find relevant information within the middle of contexts. However, these methods may alter the answering pattern of the model. For instance, Yu \cite{yu2023paraphrasing} augmented the answers by adding a paraphrase of each relevant document, sometimes leading to verbose answers that do not align with human preferences. 

 To mitigate the \textit{distraction issue}, a natural and straightforward approach is to reduce the number of input tokens, despite Long-Context LLMs theoretically supporting more tokens as input. This approach leads to retrieval-based methods, which typically retrieve the most relevant top-$k$ chunks or documents according to the query for input into LLMs. The performances of retrieval-based methods depend largely on the qualities of their retriever, and it has been manifested that Long-Context LLMs and retrieval-based methods have the potential to be combined to leverage the strengths of both \cite{xu2023retrieval}.  However, in our early exploration (as shown in Section \ref{distractor} and Table \ref{oracle}), we discover that a retriever with high or even $100\%$ recall does not guarantee good performance in the downstream QA tasks. The low precision of retriever, which implies the inclusion of numerous irrelevant documents (referred as \textit{distractors}) as input, can also lead to a significant degradation in QA performance.

\begin{table}[htbp]
\centering
\caption{F1 scores of various input documents combinations on LongBench English multi-document QA datasets. }
\resizebox{\linewidth}{!}{
\begin{tabular}{cccccccc}
\toprule
                     & \multicolumn{2}{c}{Input}   & \multicolumn{2}{c}{$\text{Retriever}$} & \multirow{2}{*}{HQA} & \multirow{2}{*}{2WIKI} & \multirow{2}{*}{MSQ} \\
\multicolumn{1}{c}{} & $\text{Pos}$          & $\text{Neg}$          & \text{Recall}       & \text{Precision}    &                           &                           &                          \\
\midrule
& $\checkmark$ & $\checkmark$ & $100\%$          & $\text{low}$            & 54.79                     & 51.03                     & 35.54                    \\
& $\checkmark$ & $\times$     & $100\%$          & $100\%$            & \textbf{64.05}                     & \textbf{65.40}                     & \textbf{52.31}                    \\
& $\times$     & $\checkmark$ & $0\%$            & $0\%$              & 26.82                     & 22.54                     & 8.36                    \\                    \bottomrule
%\multicolumn{8}{l}{\small $1.$ The base model is ChatGLM3-6B-32K.}\\ 
%\multicolumn{8}{l}{\small $2.$ Pos: a set of relevant documents, Neg: a set of irrelevant documents.}\\ 
%\multicolumn{8}{l}{\small $3.$ Input documents can be seen as retrieved by a hypothetical retriever.}\\ 
%\multicolumn{8}{l}{\small $4.$ R: recall, P: precision.}\\ 
\end{tabular}
}

\label{oracle}
\end{table}

In light of observations and analyses mentioned above, we propose an integrated Context Filtering Language Model (FltLM) aimed at enhancing the performance of Long-Context LLMs in multi-document QA tasks. FltLM is developed from vanilla Long-Context LLM with negligible number of introduced parameters, yet it manages to perform following two subtasks in order \textit{in a single forward pass}: first, discriminating distractors and filtering them out via a soft mask mechanism; and second, comprehending or reasoning based on the remaining relevant documents to generate the answer. The former task is carried out by a context filter, while the latter is performed by the Long-Context LLM. Our main contributions are summarized as follows:
\begin{itemize}
    \item We are the first to propose a context filter that can \textit{automatically} identify all distractors based on the hidden text embedding of each document. In contrast, a typical retriever only outputs a sorted list of query-document relevance scores, requiring \textit{manual} selection of input documents \textit{case by case} using top-$k$ or top-$p$ strategies to ensure the best answer quality. The training objective of the context filter also encourages model to focus on documents at any position, thus alleviating the \textit{lost in the middle} phenomenon.

    \item To filter out distractors, we design a soft mask mechanism that allows model to dynamically mask irrelevant tokens based on the discrimination results, which helps the model concentrate on relevant tokens and therefore mitigates the \textit{distraction} issue. Moreover, this soft mask design makes the entire forward pass differentiable, enabling the joint end-to-end optimization of the context filter and the Long-Context LLM.

    \item Experimental results demonstrate that our proposed one-stage integrated FltLM significantly outperforms vanilla supervised fine-tuning and two-stage retrieval-based methods. By addressing the challenges and leveraging innovative solutions, FltLM seeks to redefine the capabilities of Long-Context LLMs in processing and understanding extensive and complex textual information.

\end{itemize}

%%%%%%%%%%%%%%%%%%%%%%%%%%%%%%%%%%%%%%%%%%%%%%%%%%%%%%%%%%%%%%%%%%%%%%%%

\begin{figure*}[htbp]
    \centering
    \begin{minipage}[htbp]{0.55\linewidth}
        \centering
        \includegraphics[width=0.95\linewidth]{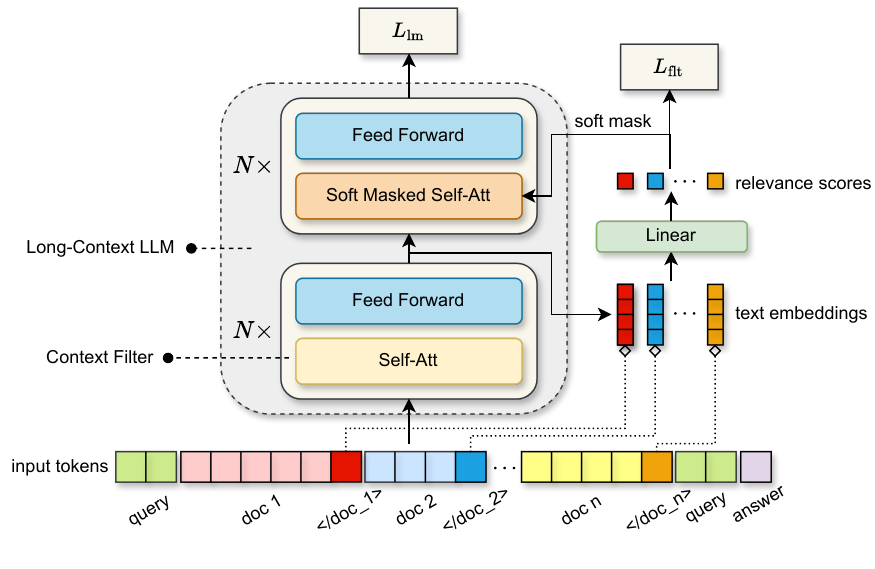}
        %\caption{The main architecture of FltLM.}
    \end{minipage}
    \begin{minipage}[htbp]{0.33\linewidth}
        \centering
        \includegraphics[width=0.95\linewidth]{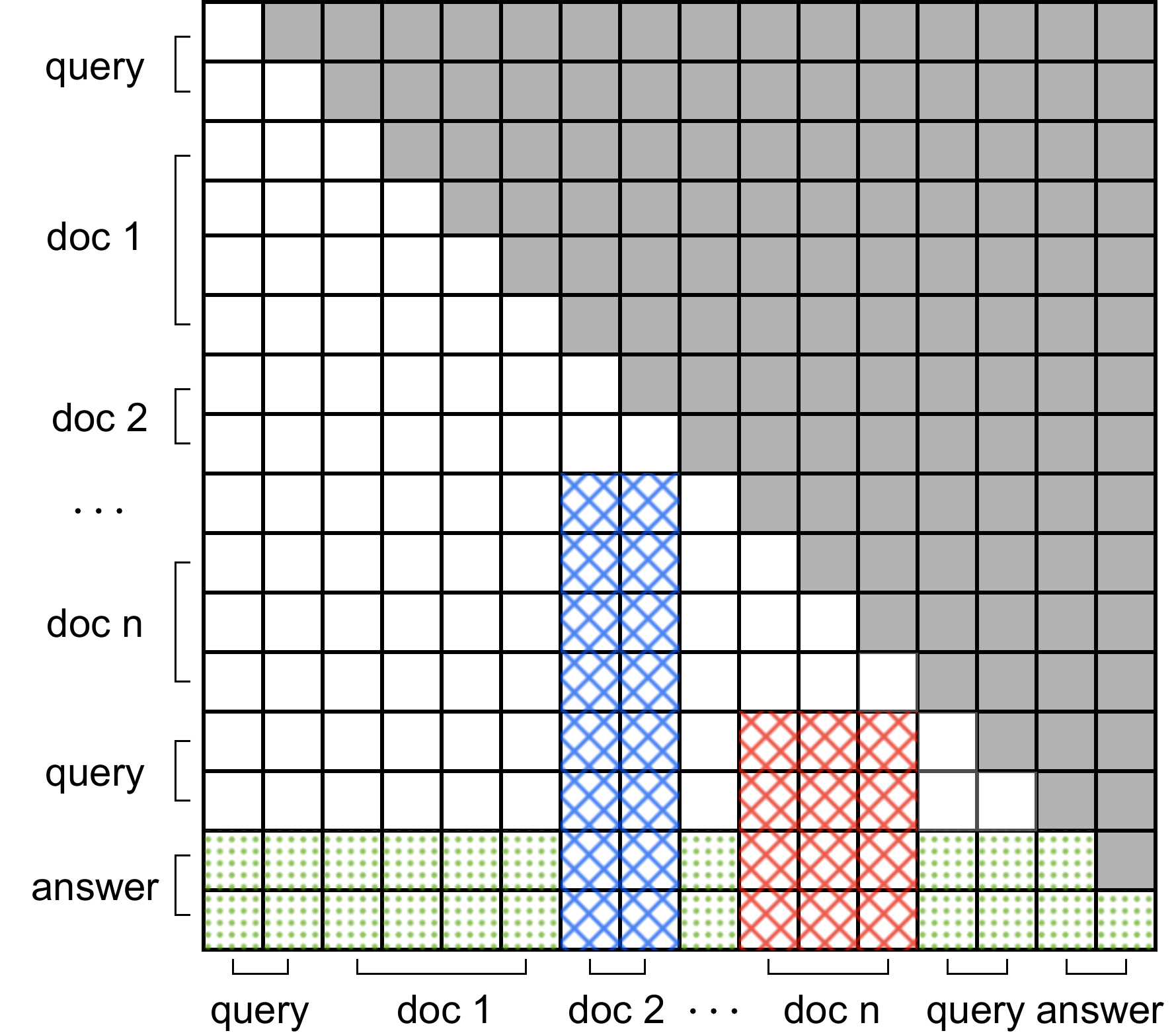}
        %\caption{An overview of the soft mask mechanism.}
    \end{minipage}
    \caption{Overview. \textbf{Left:} Main architecture of FltLM. FltLM is built upon a $2N$-layer Long-Context LLM and integrates an $N$-layer context filter designed to compute relevance scores for each document. These scores are derived from text embeddings extracted from special input tokens </doc\_i>, and are used to calculate the soft mask, which dynamically adjusts the self-attention mechanism in the last $N$ layers of the Long-Context LLM. \textbf{Right:} Soft mask mechanism.  Solid gray-filled positions signify hard masks, while cross-hatched positions represent soft masks. {\color{red}Red} and {\color{blue}blue} colors indicate different mask intensities, determined by the relevance scores from the context filter. Thanks to the soft mask mechanism, attention tends to be focused on relevant tokens during the inference stage, marked by dotted {\color{green}green} positions, to generate high-quality answers.}

    \label{fltlm}
 \end{figure*}

%\begin{figure}[htbp]
%\centering
%\includegraphics[width=0.9\linewidth]{flt.drawio-14.pdf}
%\captionsetup{justification=raggedright,singlelinecheck=false}
%\caption{The main architecture of FltLM.}
%\label{fltlm}
%\end{figure}

\section{Related work}
\subsection{Data-oriented methods}
Data-oriented methods aim to address the \textit{lost in the middle} phenomenon by constructing more informative supervising signals to strengthen the attention of the model. He et al. \cite{junqing2023never} proposed the Attention-Strengthening Multi-doc QA task, where labeled answers were organized in the order of question repetition, index prediction and answer summarization. This explicit extraction of the question and relevant document indices resembles the process of Chain-of-Thought \cite{wei2022chain}, helping the model learn the reasoning pattern comprehensively during training. Yu \cite{yu2023paraphrasing} further augmented answers by adding paraphrases of relevant documents instead of solely predicting their indices. However, these approaches alter the answering pattern of the model and may result in verbose responses, even when the prompt does not ask to do so.

In accordance with the spirit of above methods, we also leverage additional information, i.e., the index of relevant documents to enhance long-context capabilities. However, our supervising signal is not presented in natural language form but rather as a list of 0/1 labels. As a consequence, our FltLM does not alter answering habits of the model. Meanwhile, under the guidance of context filter loss (defined in Section \ref{filter}) rather than the language modeling loss, our model finds it easier to learn attention to documents at any position.

\subsection{Retrieval-based methods}
\label{retrieval}
Retrieval-based methods employ a retriever to compute relevance scores between the query and all documents. Recent research mainly focus on dense retrieval \cite{luo2024bge, chen2024bge, zhang2023retrieve, wang2023improving, ma2023fine, xiao2023c, springer2024repetition}, where a deep model learns text embeddings of query and documents and computes relevance scores with InfoNCE loss (or its variants) minimized.

 The relevance score learned by InfoNCE loss turns out to be effective. However, retrieval-based methods face two challenges: first, in practice, top-$k$ strategy is usually adopted to get retrieval results. In this process, lots of distractors may be introduced to ensure a high recall rate, which may compromise the performance of downstream multi-document QA tasks. Second, it is inadequate to determine whether a document is relevant to the query by relying on the single value of the relevance score, since InfoNCE loss is \textit{shift-invariant} (disscussed in Section \ref{filter}), and only the differences between relevance scores are meaningful. 

 In our work, we solve the above issues by modifying the InfoNCE loss, enabling our model to function as a context filter capable of identifying and filtering out all distractors.

%%%%%%%%%%%%%%%%%%%%%%%%%%%%%%%%%%%%%%%%%%%%%%%%%%%%%%%%%%%%%%%%%%%%%%%%

\section{Negative influence of distractors on multi-document QA task}
\label{distractor}
In our early exploration, we examine the impact of relevant documents and distractors on the multi-document QA task. We select all LongBench \cite{bai2023longbench} English multi-document QA datasets for experiments, using \texttt{chatglm3-6b-32k} \cite{zeng2022glm} as the Long-Context LLM, which has reported the best performance on these datasets. Three combinations of input documents are tested, including all documents (\text{Pos + Neg}), relevant documents only (\text{Pos}) and distractors only (\text{Neg}). These combinations are designed to simulate an ordinary retriever with 100\% recall and low precision, an oracle retriever with 100\% recall and precision, and an incompetent retriever with 0\% recall and precision respectively. 

Results presented in Table \ref{oracle} highlight that even when all relevant information is provided to the model, the introduction of numerous distractors in the input context significantly harms the performance on multi-document QA tasks, with degradations of $64.05\%\rightarrow54.79\%, 65.40\%\rightarrow51.03\%$ and $52.31\%\rightarrow35.54\%$ across three datasets, respectively. Additionally, we evaluate model performance in a distractors-only setting, demonstrating that the model gains certain knowledge during pre-training and can correctly answers a subset of questions even in the absence of relevant documents. However, its performance is far from comparable to settings where relevant documents are available.

\section{Methodology}
In this section, we present our FltLM in detail. Figure \ref{fltlm}(left) provides an overview of the main architecture of FltLM, wherein prompt tokens, normalization layers and residual connections have been omitted for clarity. The key insight of FltLM is straightforward: identifying all distractors and filtering them out to generate high-quality answers. 

To elaborate, we assume a Long-Context LLM has $2N$ layers. During each forward pass, we utilize the initial $N$ layers of model to identify distractors detrimental to the answer. In this process, relevance scores are computed according to the semantic text embeddings output by the $N$-th layer. Based on these scores, a soft attention mask is applied to the last $N$ layers in an adaptive manner to mask out less informative tokens, concentrating the model on relevant documents to get better answers. Thanks to its integrated design, FltLM is able to function as both a context filter and a multi-document reader effectively.

\subsection{Extraction of semantic text embeddings for Long-Context LLMs}
\label{ext}
Recently, LLM-based text embedding models are prevailing and exhibit state-of-the-art retrieval performance \cite{ma2023fine, wang2023improving, springer2024repetition}. Different from BERT-style models characterized by bi-directional attention, they typically append a </s> token to the end of the text (either query or documents), and acquire its semantic vector by extracting the embedding from the last layer of </s>. Relevance scores $s_i$ are subsequently computed as the cosine similarity between the query embedding $h_q$ and the document embedding $h_{d_i}$:
\begin{equation}
\begin{aligned}
    h_q &= \text{LLM}(q\text{</s>})[-1], \; h_{d_i} = \text{LLM}(d_i\text{</s>})[-1] \\
    s_i &= \langle h_q, h_{d_i} \rangle / \Vert h_q \Vert \Vert h_{d_i} \Vert.
\end{aligned}
\end{equation}
where $q$ and $d_i$ denotes the query and the $i$-th document respectively. For the reranker model that prioritizes accuracy over speed in ranking the input documents, relevance scores can be calculated as
\begin{equation}
\label{reranker}
\begin{aligned}
    \text{input} &=\text{"Query:}\{q\}\text{ Document:}\{d_i\}\text{</s>"} \\
    s_i &= \text{Linear}(\text{LLM}(\text{input})[-1]),
\end{aligned}
\end{equation}
where $\text{Linear}(\cdot)$ represents a linear layer utilized for regressing the relevance score based on the last layer embedding of the </s> token.

In our work, we extend Eqn. (\ref{reranker}) to encode multiple documents simultaneously, while keeping the input format of multi-document QA tasks to the largest extent. The only modification to the conventional QA input format involves appending a special token </doc\_i> following each $i$-th document:

\begin{equation}
\label{naive}
\begin{aligned}
    \text{input} =\text{"}& \{\text{prompt}\} \\
    &\text{Question:} \{q\} \\
    & \text{Document 1:} \{d_1\}\text{</doc\_1>} \\
    & \text{Document 2:} \{d_2\}\text{</doc\_2>} \\
    &\cdots \\
    &\text{Document n:} \{d_n\}\text{</doc\_n>} \\
    & \{\text{prompt}\} \\
    &\text{Question:} \{q\} \\
    &\text{Answer:\{}a\text{\}"} \\
    s_i = \text{L}&\text{inear}(\text{LLM}(\text{input})[p_i]),
\end{aligned}
\end{equation}
where $p_i$ represents the position index of </doc\_i>. During the training and inference stages, the placeholder $a$ is replaced with the labeled answer or an empty string, respectively. Additionally, we adopt query-aware contextualization \cite{liu2024lost}, which places the query before and after the documents, to construct our input. We refer Eqn.(\ref{naive}) as the \textit{naive} strategy for the extraction of text embeddings.

Compared to Eqn. (\ref{reranker}), our naive strategy enables the model to capture richer contextual information. This enhancement is achieved by exposing not only the $i$-th document but also the previous $(i-1)$ documents, along with their corresponding special tokens, to the special token </doc\_i>. Although these preceding tokens are not directly related to the $i$-th document, they contribute additional contrastive information that improves the organization of the (key, value) space of the model. This concept aligns with the approach discussed in \cite{tworkowski2024focused}. Consequently, Eqn. (\ref{naive}) yields more representative and discriminative text embeddings for each document.

In addition to the naive strategy, we explore two alternative approaches for extracting text embeddings as part of our ablation studies: 
\begin{itemize}
    \item First, we extract the embedding for each document according to Eqn. (\ref{naive}), while applying extra attention masks to the context filter to force each document to be invisible by others. In this way, text embeddings are extracted \textit{independently}.
    \item Second, we propose a setting in which </doc\_i> serves as a proxy for the aggregated information of the first $i$ documents. Under this \textit{accumulative} strategy, relevance scores are formulated as follows:
    \begin{equation}
    \label{accumulative}
    s_i = \text{Linear}(\text{LLM}(\text{input})[p_i] - \text{LLM}(\text{input})[p_{i-1}]).
\end{equation}
\end{itemize}
For detailed comparisons of these strategies, please see Section \ref{ablation}.

\subsection{Training loss of context filter}
\label{filter}
Most LLM-based retriever is trained under the guidance of the following InfoNCE loss:
\begin{equation}
\label{infonce}
\begin{aligned}
    L_{\text{InfoNCE}} &= -\log \frac{e^{s_p / \tau}}{e^{s_p / \tau} + {\sum\limits_{i\in \text{Neg}}{e^{s_i/\tau}}}} \\
    &= \log\left( 1 + \sum\limits_{i\in \text{Neg}} e^{{(s_i - s_p)/\tau}}\right) ,
\end{aligned}
\end{equation}
where $p$ is the index of positive (relevant) document and $\text{Neg}$ stands for indices of all negative (irrelevant) documents. $\tau$ is the temperature parameter. 

However, the \textit{shift-invariant} nature of InfoNCE loss dooms that relevance scores $\{s_1, s_2, \cdots, s_n\}$ share the same loss with $\{s_1 + c, s_2 + c, \cdots, s_n + c\}$, making it theoretically infeasible to establish a universal threshold value across various query-document pairs to determine whether a document is relevant to the query. To confront this challenge, we introduce an absolute threshold $s^{*}$, expecting that the learned $s_i < s^{*}$ if and only if the $i$-th document is irrelevant. We set $s^* = 0$ without loss of generality, and the InfoNCE loss can be modified by adding two regularization terms, $e^{-s_p/\tau}$ for positive document with score less than $0$, and $\sum_{i\in \text{Neg}} e^{s_i/\tau}$ for negative documents with scores greater than $0$, thus imposing large penalty on inaccurately scored documents:
\begin{equation}
\label{tmp-loss}
\begin{aligned}
    L_{\text{InfoNCE}}^{*}& \\= \log &\left( 1 + \sum\limits_{i\in \text{Neg}} e^{(s_i - s_p)/\tau} + e^{-s_p/\tau} + \sum\limits_{i\in \text{Neg}} e^{s_i/\tau} \right) \\
    = \log& \left(1 + e^{-s_p/\tau} \right) + \log \left(1 + \sum\limits_{i\in \text{Neg}} e^{s_i/\tau} \right).
\end{aligned}
\end{equation}

We then extend Eqn. (\ref{tmp-loss}) to accommodate multi-hop QA tasks, where models are asked to reading across multiple positive documents to synthesis an answer. Additionally, considering that overlooking a positive document poses greater risks than the inclusion of extra distractors, we set a margin $m > 0$ to encourage relevance scores of all positive documents exceed $m$. Ultimately, our training loss for context filter is represented as follows:
\begin{equation}
\label{our-loss}
    L_{\text{flt}} = \log \left(1 + \sum\limits_{i\in \text{Pos}} e^{-(s_i-m)/\tau} \right) + \log \left(1 + \sum\limits_{i\in \text{Neg}} e^{s_i/\tau} \right),
\end{equation}
which shares similar spirit with ZLPR loss proposed by Su et al. \cite{su2022zlpr}.

\subsection{Soft mask mechanism}
The concept of soft mask is relative to that of typical hard masks, which are implemented via adding $-\infty$ biases to the attention scores during the computation of self-attention. This hard mask operation makes specific tokens completely invisible to others. In contrast, we design a learnable soft mask mechanism to make this operation differentiable and thereby more adaptable. Specifically,  we compute mask intensities $I_i$ for each document based on their relevance scores as follows:
\begin{equation}
\label{soft-mask}
    I_i = \min\{ 0, w s_i + b \},
\end{equation}
where $w$ and $b$ are trainable parameters. We hypothesize that $w > 0$, a proposition supported by subsequent experimental results, indicating that mask intensity is positively correlated with the relevance score. We also introduce a bias $b$, allowing our model to learn to either mask less significant positive documents as well (in the case where $b < 0$), or merely mask highly significant distractors (in the case where $b > 0$).

For the last $N$ layers of the model, we augment original attention matrix $A$ by directly adding the computed intensities as follows:
\begin{equation}
    A[u_i\text{: } ,\; l_i\text{: } u_i] \mathrel{+}=  I_i,
\end{equation}
where $l_i$ and $u_i$ stand for the lower and upper index bounds of the $i$-th document, respectively. Figure \ref{fltlm}(right) illustrates our soft mask mechanism, assuming that document $2$ and $n$ are identified as distractors. In this way, we reduce the visibility of irrelevant information to subsequent tokens during answer generation, therefore enhancing the performance of multi-document QA.

\begin{table*}[htbp]
\caption{QA performance of various methods across multiple datasets. }
\label{main}
\centering
\begin{tabular}{p{2.8cm}p{1.2cm}<{\centering}p{1.2cm}<{\centering}p{1.2cm}<{\centering}p{0.8cm}<{\centering}p{0.8cm}<{\centering}p{0.8cm}<{\centering}p{0.8cm}<{\centering}}
\toprule
\multicolumn{1}{l}{\multirow{2}{*}{Methods}} & \multicolumn{3}{c}{Experimental Settings}             & \multicolumn{1}{c}{\multirow{2}{*}{HQA}} & \multicolumn{1}{c}{\multirow{2}{*}{2WIKI}} & \multicolumn{1}{c}{\multirow{2}{*}{MSQ}} & \multicolumn{1}{c}{\multirow{2}{*}{Avg.}} \\
\multicolumn{1}{l}{}                  & $L_\text{lm}$ & $\lambda L_\text{flt}$ & Soft Mask    & \multicolumn{1}{l}{}                     & \multicolumn{1}{l}{}                       & \multicolumn{1}{l}{}                     & \multicolumn{1}{l}{}                      \\
\midrule
Baseline                              & $\times$      & $\times$               & $\times$     & 54.79                                    & 51.03                                      & 35.54                                    & 47.12                                     \\
Baseline + Retrieval                              & $\times$      & $\times$               & $\times$     & 55.63                                    & 55.71                                      & 39.36                                    & 50.23                                     \\
SFT                          & $\checkmark$  & $\times$               & $\times$     & 63.72                                    & 78.73                                      & 53.28                                    & 65.24                                     \\
SFT + Retrieval                          & $\checkmark$  & $\times$               & $\times$     & 62.89                                   & 79.21                                      & 53.83                                    & 65.31                                     \\
\midrule
FltLM (w/o soft mask) & $\checkmark$  & $\checkmark$           & $\times$     & {\underline{65.67}}    & \textbf{80.39} & {\underline{54.80}}     & {\underline{66.95}}    \\
FltLM               & $\checkmark$  & $\checkmark$           & $\checkmark$ & \textbf{67.53} & {\underline{80.16}}    & \textbf{55.05} & \textbf{67.58} \\
\bottomrule
\end{tabular}
\end{table*}

\subsection{FltLM}
\label{hyper-lambda}
The training loss of FltLM is a weighted summation of following two losses: the context filter loss $L_{\text{flt}}$ supervised by indices of ground-truth documents, and language modeling loss $L_\text{lm}$ supervised by labeled answers. Formally, it can be written as
\begin{equation}
    L = L_{\text{lm}} + \lambda L_{\text{flt}},
\end{equation}
where $\lambda$ is a hyper-parameter to balance the learning of context filter and Long-Context LLM.

%%%%%%%%%%%%%%%%%%%%%%%%%%%%%%%%%%%%%%%%%%%%%%%%%%%%%%%%%%%%%%%%%%%%%%%%

\section{Experiments}
\subsection{Training data construction}
 We collect data from the training sets of following three multi-hop QA datasets: HotpotQA \cite{yang2018hotpotqa}, 2WikiMultiHopQA \cite{ho-etal-2020-constructing} and MuSiQue \cite{trivedi2021musique}, all of which are based on Wikipedia. To meet the requirements of training a Long-Context LLM that necessitates long input sequences, we replace all short paragraphs with the corresponding full articles from the Wikipedia dataset \cite{wikidump} by matching their titles. This dataset contains full Wikipedia articles that have been preprocessed to remove markdown formatting and unwanted sections. We successfully match a total of 86,882 training samples, including 19,329 for HotpotQA, 55,695 for 2WikiMultiHopQA, and 10,858 for MuSiQue. For each sample, we construct long input data using all relevant documents, and then progressively introduce distractors until the length approaches  \textasciitilde32k tokens. We shuffle the position of each document and exclude any samples that exceed the maximum input length or computational constraints of our devices. The final training set comprises 84,762 samples.

\subsection{Experimental settings}
\textbf{Training schemes}. We initialize our Long-Context LLM with the \texttt{chatglm3-6b-32k} \cite{zeng2022glm} checkpoint, renowned for its strong performance and state-of-the-art results on the LongBench \cite{bai2023longbench}, a comprehensive benchmark for long-context understanding. We choose the \textit{naive} strategy for text embeddings extraction unless specifically stated. For hyper-parameters, we set $\lambda = 0.5$ and designate $\tau$ as learnable. Our model is trained using LoRA \cite{hu2022lora} with a rank of $r = 16$, a dropout ratio of $p = 0.1$, and $\alpha = 64$, employing data-distributed parallel training \cite{li2020pytorch} with a total batch size of 32 and training epoch of 1. The maximum learning rates are set to $1 \times 10^{-4}$ for all LoRA modules and $1 \times 10^{-2}$ for $w$, $b$, and the linear layer that computes $s_i$, following a linear decay schedule with a warm-up ratio of 0.01. To reduce GPU memory usage, we apply techniques such as Flash Attention v2 \cite{dao2023flashattention}, mixed precision training \cite{micikevicius2018mixed}, and gradient checkpointing \cite{chen2016training}. All experiments can run on 4 $\times$ 80G Nvidia A800 GPUs.

\textbf{Evaluation for QA performance}. We evaluate our model using the three English multi-document QA datasets include in LongBench, specifically referred to as HQA, 2WIKI, and MSQ. These datasets are derived from the testing sets of HotpotQA, 2WikiMultiHopQA, and MuSiQue respectively, and we have verified that there is no overlap of questions between these datasets and our training set. In line with the evaluation metrics of LongBench, we use the N-gram based F1-score to measure the quality of generated answers.

\textbf{Evaluation for context filter}. We recover the ground-truth documents for HQA, 2WIKI, and MSQ by exactly matching each sample with its corresponding entry in the original datasets. Based on these matches, we evaluate the performance of our context filter through metrics such as recall, precision, and F1-score, providing a comprehensive analysis of its efficacy in filtering irrelevant contextual information.

\subsection{Main results and comparisons}

\textbf{The effectiveness of FltLM}.
To evaluate the effectiveness of our FltLM, we compare it against several mainstream solutions for multi-document QA tasks, including: (i) utilizing a general Long-Context LLM (Baseline); (ii) fine-tuning the Long-Context LLM with labeled data (SFT); and (iii) retrieval-based methods combined with the aforementioned two models. For document retrieval, we employ a state-of-the-art retriever, BGE-reranker-v2-m3 \cite{chen2024bge}, to fetch the top-$k$ documents to ensure a high recall rate of $95\%$.

Our main results, depicted in Table \ref{main}, showcase the QA performance of various methods. Notably, the best results are highlighted in \textbf{bold}, with the second best results \underline{underlined}. Examining Table \ref{main}, it is evident that our FltLM significantly outperforms supervised fine-tuning and retrieval-based methods. On average, it achieves substantial improvements of $2.34\%$ ($65.24\% \rightarrow 67.58\%$) and $2.31\%$ ($65.31\% \rightarrow 67.58\%$) respectively, and these enhancements are consistent across all datasets.

The optimized parameters, $w=0.289$ and $b=-0.206$, align with our expectation that $w>0$. This result indicates that soft masks should be applied to both irrelevant documents and marginally relevant ones.

\begin{figure*}[htbp]
    \centering
    \begin{minipage}[htbp]{0.33\linewidth}
        \centering
        \includegraphics[width=0.9\linewidth]{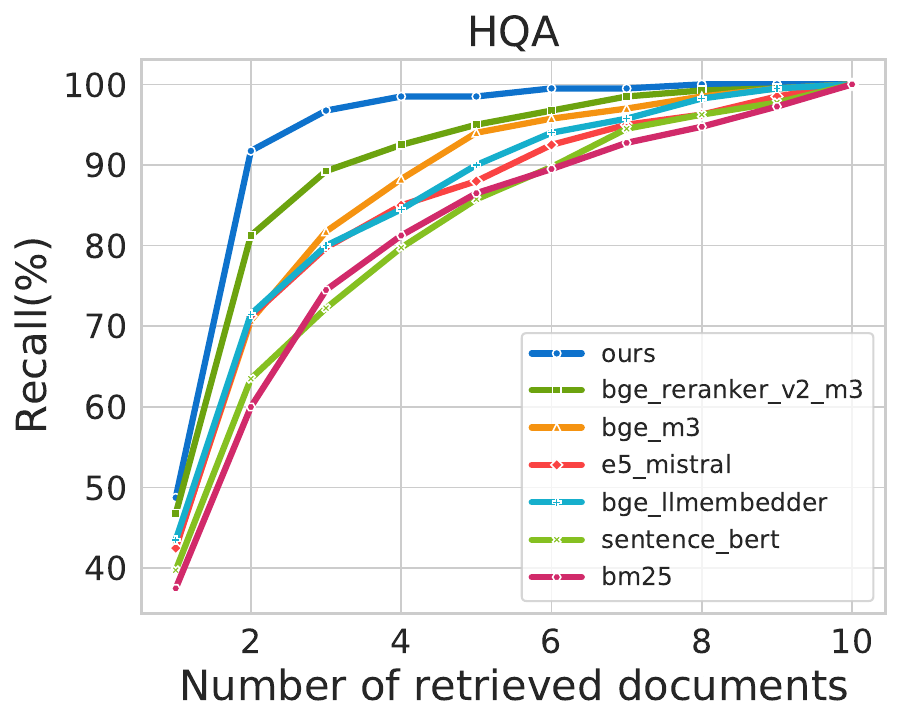}
        \subcaption{HQA dataset.}
    \end{minipage}
    \begin{minipage}[htbp]{0.33\linewidth}
        \centering
        \includegraphics[width=0.9\linewidth]{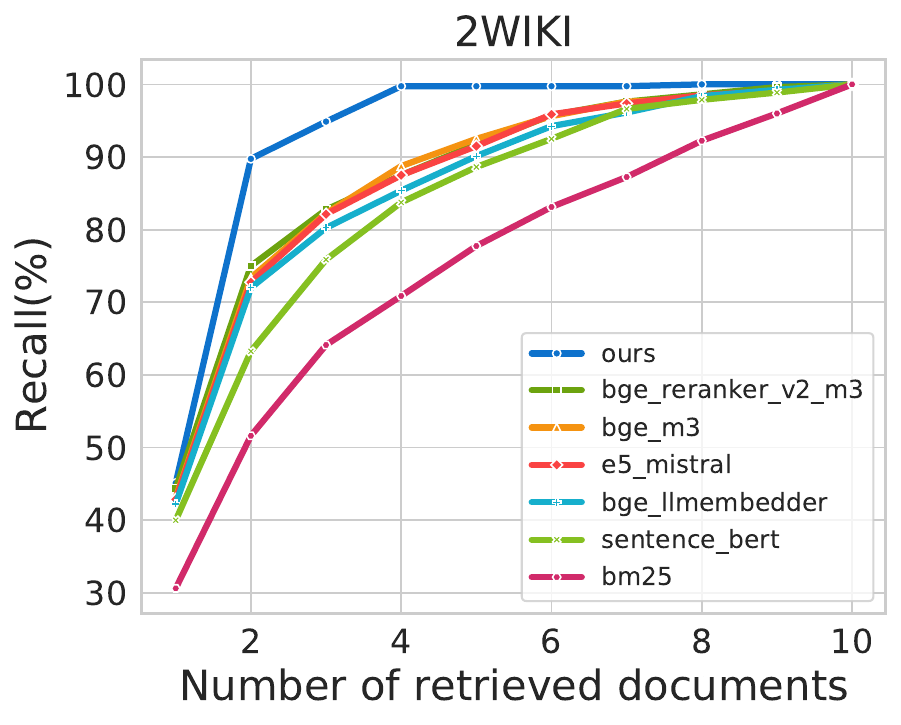}
        \subcaption{2WIKI dataset.}
    \end{minipage}
    \begin{minipage}[htbp]{0.33\linewidth}
        \centering
        \includegraphics[width=0.9\linewidth]{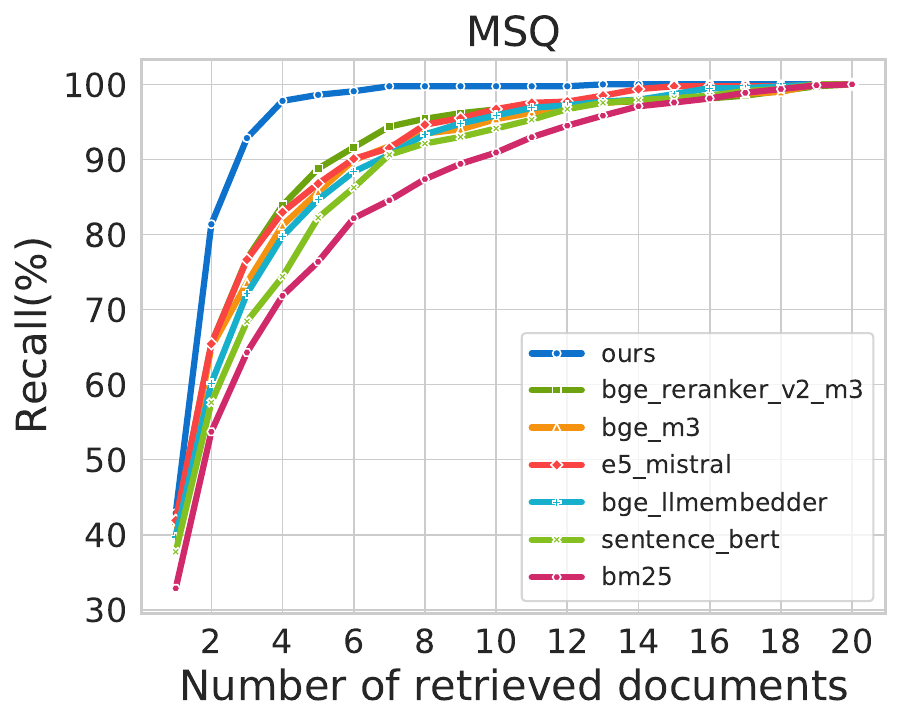}
        \subcaption{MSQ dataset.}
    \end{minipage}
    \caption{Recall of different retrievers across multiple datasets.}
    \label{recall}
 \end{figure*}

To further validate the effectiveness of each component within FltLM, we also conduct experiments on a version of FltLM devoid of the soft mask. Even without this feature, it also achieves higher F1-score compared to supervised fine-tuning ($65.24\% \rightarrow 66.95\%$). We attribute these improvements to two main factors. First, the additional term $\lambda L_{\text{flt}}$ involves with labels from ground-truth documents, providing extra supervising signals that boost our model to learn more knowledge. Second, as discussed in Section \ref{intro}, the intrinsic bias of natural language may allow Long-Context LLMs to neglect mid-text contents while still performing well on the next token prediction task. However, the distractors prediction task introduced by $\lambda L_{\text{flt}}$ requires a comprehensive understanding of all documents. Consequently, our FltLM is able to learn better semantic features that benefit the downstream tasks.

On the other hand, the incorporation of the soft mask mechanism also improves our model, evidenced by an increase in the QA F1-score of $0.63\%$ ($66.95\% \rightarrow 67.58\%$). This mechanism effectively filters out distractors, allowing our FltLM to concentrate more on relevant information and thereby mitigating the \textit{distraction} issue. Furthermore, we highlight that this modification is risk-free since $w$ and $b$ are learnable. Setting $w=b=0$ and keeping them fixed can directly degrade FltLM to its variant that lacks the soft mask mechanism.

\begin{table}[htbp]
\centering
\caption{Impact of input document order on QA performance.}
\label{lost-in-the-middle}
\begin{tabular}{cccccc}
\toprule
                          & Doc. Order     & HQA   & 2WIKI & MSQ   & Avg.          \\
\midrule
\multirow{2}{*}{Baseline} & original  & 54.79 & 51.03 & 35.54 & 47.12         \\
                          & reordered & 58.16 & 52.27 & 42.11 & 50.85 (+3.73) \\
                          \midrule
\multirow{2}{*}{SFT}      & original  & 63.72 & 78.73 & 53.28 & 65.24         \\
                          & reordered & 65.09 & 80.78 & 57.25 & 67.71 (+2.46) \\
                          \midrule
\multirow{2}{*}{FltLM}    & original  & 67.53 & 80.16 & 55.05 & 67.58         \\
                          & reordered & 67.31 & 79.26 & 59.46 & 68.68 (\textbf{+1.10}) \\
                          \bottomrule
\end{tabular}
\end{table}

\textbf{Analysis of the \textit{lost in the middle} phenomenon}.
To further confirm that FltLM alleviates the lost in the middle phenomenon, we conduct experiments by reordering the input documents. Specifically, we move half of the relevant documents to the beginning of the input and the other half to the end, while maintaining relative positions of the remaining documents. If the lost in the middle phenomenon does not exist, this reordering will not affect the model’s performance. However, as shown in Table \ref{lost-in-the-middle}, the average F1-score of the baseline model increases by $3.73\%$ ($47.12\% \rightarrow 50.85\%$), highlighting a severe lost in the middle problem. In contrast, the average F1-score of the SFT model increases by $2.46\%$ ($65.24\% \rightarrow 67.71\%$), while that of FltLM increases by only $1.10\%$ ($67.58\% \rightarrow 68.68\%$). These results suggest that our model effectively strengthens attentions to the middle relevant, making it more robust to document order compared to the SFT model.

\begin{table}[htbp]
\centering
\caption{Attention scores for positive and negative documents, averaged across samples ($\times 10^{-2}$).}
\label{attn}
\begin{tabular}{ccc}
\toprule
      & Positive Doc. & Negative Doc. \\ \midrule
      SFT & 2.61          & 1.19       \\
FltLM   & 3.54          & 0.31          \\
 \bottomrule  
\end{tabular}
\end{table}

\textbf{Analysis of the \textit{distraction} issue}.
To evaluate how effectively FltLM addresses the issue of distraction, we analyze the model's attention scores during the generation of the first token of the answer. The token-level attention scores are calculated by averaging across multiple attention heads in the final $N$ layers. Document-level scores are then obtained by summing the relevant token-level scores. Table \ref{attn} presents the average attention scores for both positive and negative documents. The results indicate that FltLM, enhanced by the soft mask mechanism, is more focused on relevant contents compared to the SFT model.

\textbf{The effectiveness of context filter}.
As an ancillary benefit, in the training process of FltLM, we yield a context filter designed to identify all distractors within the long input context. It is noteworthy that this context filter can also function as a conventional dense retriever by generating a sorted list of relevance scores. This capability prompts a natural question: how effectively can our context filter perform retrieval tasks? To explore this, we benchmark it against several existing retrievers, including BM25, Sentence-BERT \cite{reimers-gurevych-2019-sentence}, and BGE-llmembedder \cite{zhang2023retrieve}, as well as three state-of-the-art ones: BGE-m3 \cite{chen2024bge}, BGE-reranker-v2-m3 \cite{chen2024bge}, and E5-Mistral \cite{wang2023improving}. Figure \ref{recall} illustrates how recall varies with the number of documents retained for different retrievers. It is universally observed across different datasets that our context filter surpasses all the aforementioned retrievers, achieving the saturation of recall at the fastest rate.

\section{Ablation studies}
\label{ablation}

\textbf{FltLM v.s. two-stage filter-and-then-read strategy}. As an integrated model, FltLM can perform distractors identification and QA tasks through a single forward pass, while achieving strong performance. However, is this integrated end-to-end design necessary? To answer this question, we propose and evaluate a two-stage filter-and-then-read strategy as a comparison. Specifically, we train a context filter and concurrently fine-tune the Long-Context LLM under the guidance of $L_\text{flt}$ and $L_\text{lm}$, respectively. During the inference stage, the context filter first calculates relevance scores, and documents with $s_i > 0$ are then selected and fed into the fine-tuned Long-Context LLM for further processing.

\begin{table}[htbp]
\centering
\caption{FltLM v.s. filter-and-then-read strategy.}
\label{two-stage}
\begin{tabular}{lcccc}
\toprule
        Methods               & HQA            & 2WIKI          & MSQ            & Avg.           \\
                       \midrule
\multicolumn{1}{l}{SFT} & 63.72          & 78.73          & {\underline{53.28}}    & {\underline{65.24}}    \\
\multicolumn{1}{l}{Filter-and-then-read}   & {\underline{64.48}}    & {\underline{79.09}}    & 46.99          & 63.52          \\
\multicolumn{1}{l}{FltLM}                  & \textbf{67.53} & \textbf{80.16} & \textbf{55.05} & \textbf{67.58} \\
\bottomrule
\end{tabular}
\end{table}

\begin{table*}[htbp]
\centering
\caption{Impact of different text embedding extraction strategies on QA performance. Our naive extraction strategy results in minimal QA performance degradations, largely maintaining original capabilities of the Long-Context LLM.}
\label{impact}
\begin{tabular}{p{2.8cm}p{1.2cm}<{\centering}p{1.2cm}<{\centering}p{0.8cm}<{\centering}p{0.8cm}<{\centering}p{0.8cm}<{\centering}p{0.8cm}<{\centering}}
\toprule
\multirow{2}{*}{Extraction Strategies} & \multicolumn{2}{c}{Experimental Settings} & \multirow{2}{*}{HQA} & \multirow{2}{*}{2WIKI} & \multirow{2}{*}{MSQ} & \multirow{2}{*}{Avg.} \\
                                     & $L_\text{lm}$       & $L_\text{flt}$      &                      &                        &                      &                       \\
                                     \midrule
\multicolumn{1}{l}{None (Baseline)}                             & $\times$            & $\times$            & \textbf{54.79}       & \textbf{51.03}         & \underline{35.54}                & \textbf{47.12}        \\
\midrule
\multicolumn{1}{l}{Independent}                          & $\times$            & $\checkmark$        & 47.96                & 40.35                  & 32.32                & 40.21                 \\
\multicolumn{1}{l}{Accumulative}                         & $\times$            & $\checkmark$        & 20.34                & 5.65                   & 12.67                & 12.89           \\
\multicolumn{1}{l}{Naive (Ours)}                               & $\times$            & $\checkmark$        & \underline{50.80}                & \underline{49.11}                  & \textbf{35.98}       & \underline{45.30}                 \\
\bottomrule
\end{tabular}
\end{table*}

Table \ref{two-stage} provides evaluation results for this two-stage strategy, revealing that it is an intuitive yet less effective method, with an average decline of $-4.06\%$ ($67.58\% \rightarrow 63.52\%$) compared to the one-stage FltLM. We attribute this performance degradation to two main factors. First, although our context filter shows potential, it remains imperfect and may fail to retrieve all related information to answer the question. For instance, in the HQA and 2WIKI datasets, where the context filter performs relatively well, this strategy does improve answer quality compared to supervised fine-tuning. However, in the MSQ dataset, where the context filter exhibits poor recall and F1-score, the model frequently fails to collect sufficient relevant contexts, resulting in suboptimal answers. Moreover, the separate training of context filter and Long-Context LLM prevents the potential reciprocal benefits of combining their respective loss functions, a topic we will discuss at the end of this section.

\textbf{Comparisons of different text embeddings extraction strategies}. In our study, we adopt the \textit{naive} approach to derive text embeddings for the $i$-th document, specifically by extracting the hidden vector of </doc\_i>, similar to BGE-landmark \cite{luo2024bge}. However, this method presents a potential issue since previous documents are also associated with this special token, raising doubts on its adequacy in capturing the unique semantic features of the $i$-th document. To address this concern and validate our approach, we also implement two alternative strategies introduced in section \ref{ext} to train our context filter. Meanwhile, as a baseline, we experiment with a \textit{pairwise} training strategy as well, where each forward pass computes relevance score for a single document using Eqn. (\ref{reranker}).

\begin{table}[htbp]
\centering
\caption{Comparison of context filter metrics across different text embedding extraction strategies.}
\label{extraction}
\begin{tabular}{lccc}
\toprule
   Extraction Strategies          & Precision      & Recall         & F1-score             \\
             \midrule
             \multicolumn{1}{l}{Pairwise}     & 91.42          & 83.70          & 85.35 \\

\multicolumn{1}{l}{Independent}  & 89.76          & 83.93          & 84.61          \\
\multicolumn{1}{l}{Accumulative} & \underline{93.15}          & \underline{86.86}          & \underline{88.30}    \\
\multicolumn{1}{l}{Naive (Ours)}       & \textbf{93.51} & \textbf{88.04} & \textbf{89.09} \\
\bottomrule
\end{tabular}
\end{table}

Table \ref{extraction} summarizes the average context filter metrics of various methods we explored. Our initial naive approach turns out to be the most effective, followed by the accumulative strategy. In contrast, strategy that applies hard masks and extracts embeddings independently tend to yield slightly poorer outcomes compared to the pairwise training baseline. Notably, strategies that are capable of capturing richer contextual information, namely the naive and accumulative ones, significantly outperform their counterparts, underscoring their ability to produce highly representative and discriminative text embeddings for each document. These results could provide valuable insights and potentially inspire improvements in text embedding models.

We further examine the compatibility of various text embedding extraction strategies with our ultimate goal of multi-document QA. To this end, we freeze the last $N$ layers of the Long-Context LLM and fine-tune it solely under the guidance of $L_\text{flt}$. Our findings, presented in Table \ref{impact}, reveal that the application of the naive strategy results in only minimal reductions in QA performance even without the guidance of $L_\text{lm}$. This observation suggests that the latent features of the LLM could potentially perform additional tasks beyond next token prediction, while largely retaining its original functionalities.

\textbf{The impact of $\lambda$ on the performance of FltLM}. As discussed in section \ref{hyper-lambda}, we introduce the hyper-parameter $\lambda$ to control the trade-off between the learning processes of the context filter and the Long-Context LLM. In pursuit of our objective to enhance the capabilities Long-Context LLM, we treat $L_\text{flt}$ as an auxiliary loss and roughly set $\lambda = 0.5$. To further analyze the impact of $\lambda$ on the performance of FltLM, we conduct ablation studies with $\lambda = 0.2$ and $1.0$.

\begin{figure}[htbp]
    \centering
    \includegraphics[width=0.6\linewidth]{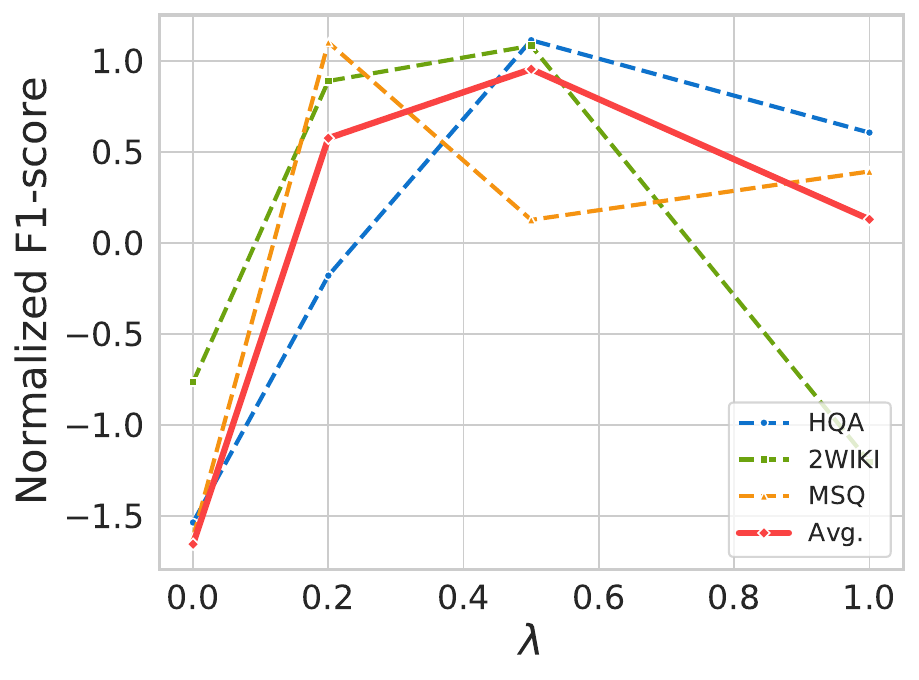}
    \caption{QA performance of FltLM across different values of $\lambda$.}
    \label{lambda}
 \end{figure}

 Figure \ref{lambda} describes how QA performance vaires with the value of $\lambda$, where $\lambda=0$ represents standard supervised fine-tuning. Metrics for each dataset are normalized independently to visualize the variation trends on the same scale. Figure \ref{lambda} reveals that FltLM consistently achieves higher F1-scores than supervised fine-tuning at $\lambda=0.2$ and $0.5$, highlighting the robustness of our method. However, performance decreases when $\lambda$ is raised to $1.0$. An optimal balance is achieved at $\lambda = 0.5$, yielding the best average results for our FltLM.

%%%%%%%%%%%%%%%%%%%%%%%%%%%%%%%%%%%%%%%%%%%%%%%%%%%%%%%%%%%%%%%%%%%%%%%%

\section{Conclusion}
In this paper, we propose FltLM, a novel integrated Long-Context LLM that significantly enhances multi-document QA performance, addressing the two critical challenges of \textit{the lost in the middle phenomenon} and the \textit{distraction} issue. FltLM employs a context filter with a soft mask mechanism which identifies and dynamically excludes the less relevant content, thereby focusing on the essential information for improved long-context understanding. By embedding the context filter directly within the architecture of the model, FltLM not only streamlines computational processes to a single forward pass but also markedly surpasses supervised fine-tuning and retrieval-based methods in complex QA settings.

The emergence of FltLM opens up new avenues for advanced natural language processing applications. Future work will focus on optimizing the context filtering process, extending the applicability of the model to other natural language processing tasks such as sophisticated document summarization and in-depth content generation, and integrating emerging neural network paradigms to further enhance performance and scalability. This progression promises to improve the capabilities of Long-Context LLMs significantly, making them more versatile and effective across various domains.

%%%%%%%%%%%%%%%%%%%%%%%%%%%%%%%%%%%%%%%%%%%%%%%%%%%%%%%%%%%%%%%%%%%%%%%%

%%% Use this environment to include acknowledgements (optional).
%%% This will be omitted in doubleblind mode.

\begin{ack}
This work was supported by the Natural Science Foundation of China under grant 62071171 and the high-performance computing platform of Peking University.
\end{ack}

%%%%%%%%%%%%%%%%%%%%%%%%%%%%%%%%%%%%%%%%%%%%%%%%%%%%%%%%%%%%%%%%%%%%%%%%

%%% Use this command to include your bibliography file.
\bibliography{mybibfile}

\section*{Appendix}

\subsection*{A. Implementation details}

\begin{table}[htbp]
\centering
\caption{Inplementation details of FltLM.}
\begin{tabular}{lc}
\toprule
& settings \\
\midrule
%Training code & based on LlamaFactory\cite{zheng2024llamafactory} \\
Initialization                      & \texttt{chatglm3-6b-32k} \\
Extraction strategy & \textit{naive}                      \\
Linear layer for context filter  & learnable \\
Soft mask parameters $w, b$ & learnable \\
Temperature $\tau$                              & learnable          \\
Margin $m$                          & learnable    \\
Hyper-parameter $\lambda$                           & 0.5                                                  \\
Special tokens embeddings & fixed to be zeros \\
LoRA rank $r$                       & 16                                                   \\
LoRA $\alpha$                       & 64                                                   \\
LoRA dropout ratio $p$              & 0.1                                                  \\
Deepspeed stage & 0 (DDP) \\
Total batch size                    & 32                                                   \\
Epoch                               & 1                                                    \\
Learning rate schedule              & linear                                               \\
Maximum learning rate               & 1e-4/1e-2 for LoRA/others               \\
Warmup ratio                        & 0.01                                                 \\
Devices & 4 $\times$ 80G Nvidia A800 GPUs \\
\bottomrule
\end{tabular}
\label{impl}
\end{table}

Table \ref{impl} outlines our implementation details. Given that both $\tau$ and $m$, as well as the linear layer for the context filter, are learnable parameters in our experiments, we can equivalently set $\tau=1$ as a fixed value while maintaining $m$ and the linear layer to be learnable. $w$ is initialized to a small positive number of $1\times10^{-3}$ and $b$ is initialized to $0$. To reduce GPU memory usage, we apply techniques such as Flash Attention v2 \cite{dao2023flashattention}, mixed precision training \cite{micikevicius2018mixed}, and gradient checkpointing \cite{chen2016training}.

\subsection*{B. Visualization of $w$ and $b$}
\begin{figure}[htbp]
    \centering
    \includegraphics[width=0.6\linewidth]{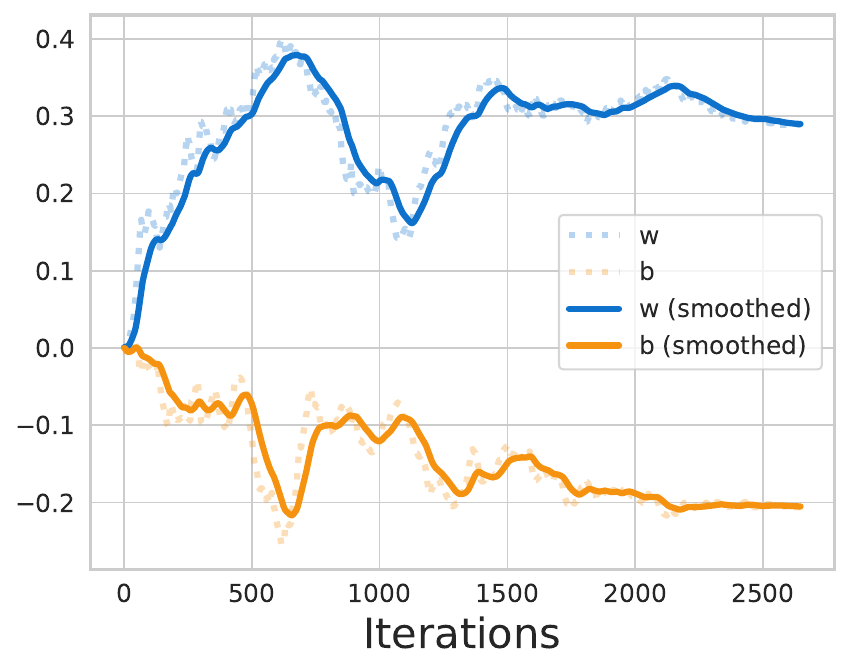}
    \caption{Evolutions of $w$ and $b$ over iterations.}
    \label{w_b}
 \end{figure}

Figure \ref{w_b} shows how $w$ and $b$ evolve over iterations.

\subsection*{C. Detailed evaluation metrics for context filter}
Table \ref{context-filter} presents the detailed evaluation metrics for this filter, indicating satisfactory performance with an average F1-score of $89.62\%$.

\begin{table}[htbp]
\centering
\caption{Evaluation metrics of the learned context filter.}
\label{context-filter}
\begin{tabular}{lcccc}
\toprule
          & HQA   & 2WIKI & MSQ   & Avg.  \\
          \midrule
Precision & 93.17 & 98.50 & 88.94 & 93.54 \\
Recall    & 88.75 & 96.00 & 82.07 & 88.94 \\
F1-score  & 89.32 & 96.42 & 83.12 & 89.62 \\
\bottomrule
\end{tabular}
\end{table}

\subsection*{D. Demonstrated case}

Here is a demonstrated case:
\begin{itemize}
    \item \textbf{Question}: During the war in which The Things They Carried is set, when was conscription introduced by the country where the film Grievous Bodily Harm was later released?
    \item \textbf{Relative contexts (simplified)}:
    
    (The Things They Carried, doc\#1) The Things They Carried is a story about American soldiers in the Vietnam War.

    (Grievous Bodily Harm, doc\#5) Grievous Bodily Harm is an Australian crime film.
    
    (Conscription in Australia, doc\#9) Vietnam War. In 1964, compulsory national service for 20-year-old males was introduced under the National Service Act 1964.

    \item \textbf{Relevance scores computed by FltLM}: [2.625, -4.46875, -6.6875, -6.3125, -7.125, 7.84375, -12.1875, -8.9375, 3.65625, -11.3125], with the score of 1st, 6th, and 9th documents > 0 (correct).

    \item \textbf{Answer of FltLM}: 1964 (correct)
    
\textbf{Answer of SFT}: 1911 (wrong)
\end{itemize}

\subsection*{E. Optimal number of layers for the context filter}

In our work, we roughly assign $N$ out of $2N$ (with a proportion of 1/2) layers for our context filter. This is inspired by an intuition that the latent features of layers close to the input or output are more aligned with the token space, whereas the features of the middle layers are more closely associated with the semantic space. Results presented in Table \ref{layer} further support our assumption, demonstrating optimal outcomes when the proportion is set at 1/2.

\begin{table}[bthp]
\centering
\caption{Impact of context filter layer proportion on QA performance.}
\begin{tabular}{lcccc}
\toprule
Proportion & HQA            & 2WIKI          & MSQ            & Avg.           \\
\midrule
1/4                          & \underline{66.52}          & 78.72          & 50.89          & 65.38          \\
1/2 (Ours)                          & \textbf{67.53} & \textbf{80.16} & \underline{55.05}          & \textbf{67.58} \\
3/4                          & 64.12          & \underline{79.03}          & \textbf{57.23} & \underline{66.79}       \\
\bottomrule
\end{tabular}
\label{layer}
\end{table}

\subsection*{F. The selection of margin $m$}

Our context filter loss is defined as

\begin{equation}
\label{our-loss}
    L_{\text{flt}} = \log \left(1 + \sum\limits_{i\in \text{Pos}} e^{-(s_i-m)/\tau} \right) + \log \left(1 + \sum\limits_{i\in \text{Neg}} e^{s_i/\tau} \right).
\end{equation}
 Although it increases monotonically with $m$, we believe a margin $m > 0$ may help context filter to learn discriminative text embeddings, which could implicitly minimize $L_\text{lm}$. Therefore, in our work, we set $m=\exp({\gamma}) > 0$ where $\gamma$ is learnable with an initial value of $0$. Experimental results show that the learned $m=0.08$. We have also tried for other values of $m$, and the outcomes are exhibited in Table \ref{margin}.

\begin{table}[htbp]
\centering
\caption{Impact of margin $m$ on QA performance.}
\begin{tabular}{ccccc}
\toprule
$m$          & HQA            & 2WIKI          & MSQ            & Avg.                      \\
\midrule
Learnable (Ours)    & \textbf{67.53} & \underline{80.16}          & \textbf{55.05} & \textbf{67.58}            \\
Fixed to be 0.5 & \underline{66.08}          & \textbf{81.43} & \underline{53.09}          & \underline{66.87}                     \\
Fixed to be 1.0 & 65.91          & 79.89          & 52.78          & {66.19} \\
\bottomrule
\end{tabular}
\label{margin}
\end{table}

\subsection*{G. Feedback from downstream tasks inversely enhance the context filter} 

A digressive yet insightful question: can feedback from downstream tasks inversely enhance the context filter? To explore this interesting hypothesis, we design a comparative experiment in which the context filter is trained under two conditions: (i) solely with the loss function $L_\text{flt}$, and (ii) with the combined loss function $L_\text{flt} + \mu L_\text{lm}$ where $\mu=0.5$. The results, as presented in Table \ref{feedback}, indicate that feedback from downstream tasks significantly boost the efficacy of the context filter. Combine with our main results of FltLM, we speculate that the loss functions $L_\text{flt}$ and $L_\text{lm}$ exhibit a reciprocal effect, mutually enhancing each other in our experimental setup.

\begin{table}[htbp]
\centering
\caption{Effect of feedback from downstream tasks on context filter.}
\label{feedback}
\begin{tabular}{lccccc}
\toprule
\multicolumn{1}{l}{}                        & Loss Function                  & HQA            & 2WIKI          & MSQ            & Avg.           \\
\midrule
\multirow{2}{*}{Precision}                  & $L_\text{flt}$ & 93.29          & \textbf{98.75} & 88.50          & 93.51          \\
                                            & $L_\text{flt}+\mu L_\text{lm}$             & \textbf{93.71} & 98.58          & \textbf{89.64} & \textbf{93.98} \\
                                            \midrule
\multicolumn{1}{l}{\multirow{2}{*}{Recall}} & $L_\text{flt}$ & 87.75          & 96.25          & 80.13          & 88.04          \\
\multicolumn{1}{l}{}                        & $L_\text{flt}+\mu L_\text{lm}$             & \textbf{88.25} & \textbf{97.00} & \textbf{81.57} & \textbf{88.94} \\
\midrule
\multirow{2}{*}{F1-score}                   & $L_\text{flt}$ & 88.92          & 96.82          & 81.53          & 89.09          \\
                                            & $L_\text{flt}+\mu L_\text{lm}$             & \textbf{89.08} & \textbf{97.21} & \textbf{83.11} & \textbf{89.80} \\
                                            \bottomrule
\end{tabular}
\end{table}

\end{document}